\documentclass{article}

\usepackage{arxiv}

\usepackage[utf8]{inputenc} 
\usepackage[T1]{fontenc}    
\usepackage{hyperref}       
\usepackage{url}            
\usepackage{booktabs}       
\usepackage{amsfonts}       
\usepackage{nicefrac}       
\usepackage{microtype}      
\usepackage{lipsum}

\usepackage{amsmath,amssymb,amsfonts}
\usepackage{algorithmic}
\usepackage{graphicx}
\usepackage{textcomp}
\usepackage[caption=false]{subfig}
\usepackage{caption,setspace}
\usepackage{textcomp}
\usepackage{array}
\usepackage{multirow}
\usepackage{multicol}
\usepackage{newverbs}
\usepackage{soul}
\usepackage{float}
\usepackage[ruled]{algorithm2e}

\renewcommand\hl[1]{1}  

\title{CNN2Gate: Toward Designing a General Framework for Implementation of Convolutional Neural Networks on FPGA}

\author{
  Alireza Ghaffari, Yvon Savaria\\
  Polytechnique Montreal \\ 
  2900 Edouard Montpetit Blvd\\
  Montreal, QC H3T 1J4, Canada.\\
  \texttt{seyed-alireza.ghaffari@polymtl.ca}}

\begin{document}
\maketitle
\begin{abstract}
Convolutional Neural Networks (CNNs) have a major impact on our society because of the numerous services they provide. These services include, but are not limited to image classification, video analysis, and speech recognition. On the other hand, they require considerable computing power. To satisfy these requirements, it is possible to use graphic processing units (GPUs). However, high power consumption and limited external IOs constrain their usability and suitability in industrial and mission-critical scenarios. Recently, the number of researches that utilize FPGAs to implement CNNs are increasing rapidly. This is due to the lower power consumption and easy reconfigurability offered by these platforms. Because of the research efforts put into topics such as architecture, synthesis and optimization, some new challenges are arising to integrate such hardware solutions to high-level machine learning software libraries. This paper introduces an integrated framework (CNN2Gate) that supports compilation of a CNN model for an FPGA target. CNN2Gate exploits the OpenCL\textsuperscript{TM} synthesis workflow for FPGAs offered by commercial vendors. CNN2Gate is capable of parsing CNN models from several popular high-level machine learning libraries such as Keras, Pytorch, Caffe2 etc. CNN2Gate extracts computation flow of layers, in addition to weights and biases and applies a ``given'' fixed-point quantization. Furthermore, it writes this information in the proper format for OpenCL synthesis tools that are then used to build and run the project on FPGA. CNN2Gate performs design-space exploration using a reinforcement learning agent and fits the design on different FPGAs with limited logic resources automatically. This paper reports results of automatic synthesis and design-space exploration of AlexNet and VGG-16 on various Intel FPGA platforms. CNN2Gate achieves a latency of 205 ms for VGG-16 and 18 ms for AlexNet on the FPGA.
\end{abstract}


\section{Introduction}
\label{sec:introduction}
The impact of machine learning and deep learning is rapidly growing in our society due to their diverse technological advantages. Convolutional neural networks (CNNs) are among the most notable architectures that provide a very powerful tool for many applications such as video and image analysis, speech recognition and recommender systems\cite{cnn_mit_rev}. On the other hand, CNNs require considerable computing power. In order to better satisfy some given requirements, it is possible to use high-performance processors like graphic processing units (GPUs)\cite{strigl2010performance}. However, GPUs have some shortcomings that limit their usability and suitability in day-to-day industrial and mission-critical scenarios. The first downside of using GPUs is their high power consumption. This makes GPUs hard to use in robotics, drones, self-driving cars and Internet of Things (IoTs) while these fields can highly benefit from deep learning algorithms. The second downside is the lack of external Inputs/Outputs (I/Os). GPUs are typically accessible through some PCI-express bus on their host computer. This makes it hard to use them in mission-critical scenarios that need prompt control actions.mission-critical scenarios that need prompt control actions.

In particular, the floating-point number representation of CPUs and GPUs is used for most deep learning algorithms. However, it is possible to benefit from custom fixed-point numbers (quantized numbers) to reduce the power consumption, circuit footprint and to increase the number of compute engines \cite{krishnamoorthi2018quantiziation}. It was proven that many convolutional neural networks can work with 8-bit quantized numbers or less\cite{8bitq}. Hence, GPUs waste significant computing power and power consumption when performing inference in most deep learning algorithms. A better balance can be restored by designing application specific computing units using quantized arithmetic. Quantized numbers provide other benefits such as memory access performance improvements as they allocate less space in the external or internal memory of compute devices. Memory access performance and memory footprint are particularly important for hardware designers in many low-power devices. Most of the portable hardware devices such as cell-phones, drones and internet of things (IoT) sensors require memory optimization because of the limited resources available on these devices.

Field Programmable Gate Arrays (FPGA) can be used in these scenarios to tackle the problems caused by limitations of GPUs without compromising the accuracy of the algorithm. Using quantized deep learning algorithms can solve the power consumption and memory efficiency issues on FPGAs as the size of the implemented circuits shrinks with quantization. In \cite{intel2017can}  the authors reported that the theoretical peak performance of 6-bit integer matrix multiplication (GEMM) in Titan X GPU is almost 180 GOP/s/Watt, while it can be as high as 380 GOP/s/Watt for Stratix 10, and 200 GOP/s/Watt for Arria 10 FPGAs. This means high end FPGAs are able to provide comparable or even better performance per Watt than the state of the art GPUs. FPGAs are scalable and configurable. Thus, a deep convolutions neural network can be configured and scaled to be used in a much smaller FPGA in order to reduce power consumption for mobile devices. This is why this work explores means to have design-space exploration that fits a CNN to smaller FPGAs. Having massive connectivity through I/Os is natural in FPGAs. 
Furthermore, FPGAs can be flexibly optimized and tailored to perform various applications in cooperation with general-purpose processors. There are many commercialized System-on-Chip (SoC) devices that integrate both processor and FPGA fabric into a single chip ( see, for example,~\cite{intelsoc}). These SoCs are widely used on mobile devices and make the use of application-specific processors obsolete in many cases . 

There are many pieces of research that address architecture, synthesis and optimization of deep learning algorithms on FPGAs \cite{2017pipecnn,fpgaconvnet_1_2018,umuroglu2017finn,ma2017optimizing, bilaniuk2019bit}. Nonetheless, little work was done about integrating the results into a single tool. {In depth literature review in Section.}\ref{RelatedWorks}{ exposes that, tools such as} \cite{pipes2015opencl,hls4ml} {do not provide neither integrated design space exploration nor generic CNN model analyzer. In }\cite{bilaniuk2019bit, ma2017optimizing, zhang2018caffeine} { the authors specifically discuss the hardware design and not the automation of the synthesis. In} \cite{fpgaconvnet_1_2018}, {fitting the design automatically in different FPGA targets is not discussed}. This presents a major challenge as there are many degrees of freedom in designing a CNN. There is no fixed architecture for CNN as a designer can choose as many  convolution, pooling and fully connected layers as needed to reach the desired accuracy. In addition, there must be an algorithm available to manage and leverage the available resources on the FPGA such as Digital Signal Processing (DSP) units, lookup tables and registers. This includes design-space exploration algorithms that optimize resources utilization on FPGA.

A number of development environments exist for the description and architectural synthesis of digital systems \cite{vivado,Quartus}. However, they are designed for general purpose applications.  This means that there is a great opportunity to make a library which uses these hardware design environments to implement deep learning algorithms. Alternately, there are also many other development frameworks/libraries for deep learning in Python language. Some notable libraries for deep learning are Pytorch, Keras, Tensorflow and Caffe. Another challenge would be designing a synthesis tool for FPGA that can accept models produced by all the aforementioned Python libraries and many more available for machine learning developers. 

This article proposes methods to tackle research challenges related to the integration of high-level synthesis tools for convolutional neural networks on FPGAs. This paper elaborates on the following contributions:

\begin{enumerate}
\item \textbf{Generalized model analysis}
Most of the previous implementations of the CNNs on FPGA fall short in supporting as many as possible machine learning libraries. CNN2Gate benefits from a generalized  {model} transfer format called ONNX (Open Neural Network eXchange format) \cite{onnx}. 
ONNX helps hardware synthesis tool to be decoupled from the framework in which the CNN was designed. Using ONNX in CNN2Gate provides us the advantage of the ability to focus on hardware synthesis without focusing on a specific machine learning tool. CNN2Gate integrates an ONNX parser that extracts the computation data-flow as well as weights and biases from the ONNX representation {of a CNN model}. It then writes these data in a format that is more usable with hardware synthesis workflows. 
\BlankLine
\item \textbf{Automated high-level synthesis tool} 

There are hardware implementation aspects of the machine learning algorithms that are not obvious for many machine learning engineers and computer scientists. High-level synthesis is one of the solutions that deal with the tremendous demand for hardware design productivity. CNN2Gate offers a high-level synthesis workflow that can be used to implement CNN models on FPGA. {CNN2Gate is a Python library that parses, synthesize and runs a CNN model automatically. It eliminates the need for FPGA experts to manually implement the CNN model targeting FPGA hardware during early stages of the design. Having the form of a Python library, CNN2Gate can be directly exploited by machine learning developers to implement a CNN model on FPGAs}. CNN2Gate is built around an available open-source tool, notably \textit{PipeCNN} \cite{2017pipecnn} that exploits the capabilities of existing design tools to use OpenCL kernels from which high-level synthesis can be performed.  
\BlankLine
\item \textbf{Hardware-aware design-space exploration}

An important aspect of design-space exploration is trying to choose design parameters in order to achieve some desired performance prior to generation of physical design. CNN2Gate provides a design-space exploration tool that is able to manipulate the level of parallelism of the algorithm to fit the design on various FPGA platforms. The design-space exploration algorithms proposed here uses the estimation of hardware resources (e.g. DSPs, lookup tables, registers and on-chip memory) to fit the design. This tool triggers the first stage of the synthesis tool provided by the FPGA vendor and receives back the estimated hardware resource utilization. In the next step, the tool tunes the design parameters according to the resource utilization feedback and iterates again to get the new hardware resource utilization. We have used two algorithms to do design-space exploration. The first algorithm is based on brute-force to check all possible parameter values. The second algorithm is based on a reinforcement learning (RL) agent that explores the design space with a set of defined policies and actions. It will be shown that the RL agent can find best results faster than the brute-force algorithm. Advantages and disadvantages of these two exploration algorithms will be discussed in the corresponding section.

\end{enumerate}

CNN2Gate is a Python library for performing inference of CNNs on FPGA that is capable of:
\begin{itemize}
\item  Parsing CNN models. 
\item  Extracting the computation flow of each layer.
\item  Extracting weights and biases of each kernel.
\item  Applying post-training quantization values to the extracted weights and biases.
\item  Writing this information in the proper format for hardware synthesis tools.
\item  Performing design space exploration for various FPGA devices using a hardware-aware reinforcement learning algorithm.
\item Synthesizing and running the project on FPGA.
\end{itemize}

This article is organized as follows. Section \ref{RelatedWorks} reviews the related works on this topics. Section \ref{bg} reviews the most relevant background knowledge on convolutional neural networks. Section \ref{arch} elaborates on how CNN2Gate extracts the computation flow, configures the kernels and does design space exploration. Then, Section \ref{res} reports some results and compare it to other existing implementations.




\section{Related Works} \label{RelatedWorks}
A great deal of research was conducted on implementing deep neural networks on FPGAs. Among those researches \textbf{\textit{hls4ml}}\cite{hls4ml}, \textbf{\textit{fpgaConvNet}}\cite{fpgaconvnet_1_2018} and  \textbf{\textit{Caffeine}}\cite{zhang2018caffeine} are the most similar to the present paper. hls4ml is a companion compiler package for machine learning inference on FPGA. It translates open-source machine learning models into high-level synthesizable (HLS) descriptions. This package was specifically developed for an application in particle physics, with the purpose of reducing development time on FPGA. However, as stated in the status page of the project \cite{hls4mlstatu}, the package only supports Keras and Tensorflow for CNNs and the support for Pytorch is in development. {In addition, to the best of our knowledge, hls4ml does not offer design-space exploration}. {FpgaConvNet} is also an end-to-end framework for the optimized mapping of CNNs on FPGAs. FpgaConvNet uses a symmetric multi-objective algorithm to optimize {the generated design for either throughput, latency, or multi-objective criteria (e.g. throughput and latency)}. The front-end parser of fpgaConvNet can analyze models expressed in Torch and Caffe machine-learning libraries. Caffeine is also a software-hardware co-design library that directly synthesize Caffe models comprising convolutional layers and fully connected layers for FPGAs. The main differences between CNN2Gate and other cited works are in three key features. First, as explained in  Section \ref{onnx}, CNN2Gate leverages a model transfer layer (ONNX), which automatically brings the support for the vast majority of machine-learning Python libraries without bounding the user to a specific machine-learning library. Second, CNN2Gate is based on OpenCL compared to hls4ml and fpgasConvNet which are based on C++. Third, CNN2Gate proposes an FPGA fitter algorithm based on reinforcement learning.

Using existing high-level synthesis technologies, it is possible to synthesize OpenCL Single Instruction Multiple Thread (SIMT) algorithms to RTL. It is worth mentioning some notable researches on this venue. In \cite{aydonat2017opencl}, the authors provided a deep learning accelerator targeting Intel's FPGA devices based on OpenCL. This architecture was capable of maximizing data-reuse and minimizing memory accesses. The authors of \cite{suda2016throughput} presented a systematic design-space exploration methodology to maximize the throughput of an OpenCL-based FPGA accelerator for a given CNN model. They used synthesis results to empirically model the FPGA resource utilization. Similarly, in \cite{zhang2015optimizing}, the authors analyzed the throughput and memory bandwidth quantitatively to tackle the problem of design-space exploration of a CNN design targeting FPGAs. They also applied various optimization methods such as loop-tiling to reach the best performance. A CNN RTL compiler is proposed in \cite{ma2016scalable}. This compiler automatically generates sets of scalable computing primitives to form an end-to-end CNN accelerator.

Another remarkable OpenCL-based implementation of CNNs is \textbf{\textit{PipeCNN}}\cite{2017pipecnn}. PipeCNN is mainly based on the capability of currently available design tools to use OpenCL kernels in high-level synthesis. PipeCNN consists of a set of configurable OpenCL kernels to accelerate CNN and optimize memory bandwidth utilization. Data reuse and task mapping techniques have also been used in that design. Recently, in \cite{pipecnn2019abm}, the authors added a sparse convolution scheme to PipeCNN to further improve the performance in terms of throughput.Our work (CNN2Gate) follows the spirit of PipeCNN. CNN2Gate is built on top of a modified version of PipeCNN which is capable of exploiting a library of primitive kernels needed to perform inference of a CNN. However, CNN2Gate also includes means to perform automated design-space exploration and can translate CNN models provided by a wide variety of machine learning libraries.

{There are several reasons that we have chosen PipeCNN as the foundation of CNN2Gate. First, benefiting from this OpenCL model and the design methods and tools with which it is compatible, when targeting an FPGA, it is possible to fine-tune the amount of parallelism in the algorithm as explained in the Section}~\ref{fitter}. {Second, it supports the possibility of having deep pipelined design as explained in Section}~\ref{archpipes}. {Third, the library of primitive kernels can be easily adapted and re-configured based on the information extracted from a CNN model} (Section \ref{onnx}).

Lately, reinforcement learning has been used to perform automated quantization for neural networks. \textbf{\textit{HAQ}}\cite{haq} and \textbf{\textit{ReLeQ}}\cite{releq} are example of researches exploiting this technique. ReLeQ proposes a systematic approach to solve the problem of deep quantization automatically. This provides a general solution for quantization of a large variety of neural networks. Likewise, HAQ, suggests a hardware-aware automated quantization method for deep neural networks using actor-critic reinforcement learning method \cite{a3csurvay}. Inspired by these two papers, we used a reinforcement learning algorithm to control the level of parallelism in CNN2Gate OpenCL kernels.

There are some other works that are worth mentioning here as they contribute to the hardware implementation aspect of designing CNN accelerators. An overview of performance comparison of high-end FPGAs and GPU is presented in \cite{intel2017can}. \textbf{\textit{FINN}} \cite{umuroglu2017finn}, is a framework that enables mapping binarized neural network in hardware. Similarly, \cite{zhao2017accelerating} also explores means to implement accelerator for binary neural networks. In \cite{bilaniuk2019bit}, a new bit-slicing matrix-vector processing unit is proposed. Optimizing loop operations and dataflow is explored in \cite{ma2017optimizing}. A fusion architecture that is able to fuse multiple layers and reuse data was explored in \cite{xiao2017exploring}.

\section{Background}\label{bg}

This section provides the background knowledge required to understand CNNs and OpenCL-based high-level synthesis workflows.
\subsection {\textbf{Convolutional Neural Networks (CNN\lowercase{s})}}

\begin{figure*}[!t]
  \centering
  \includegraphics[width=0.8\linewidth]{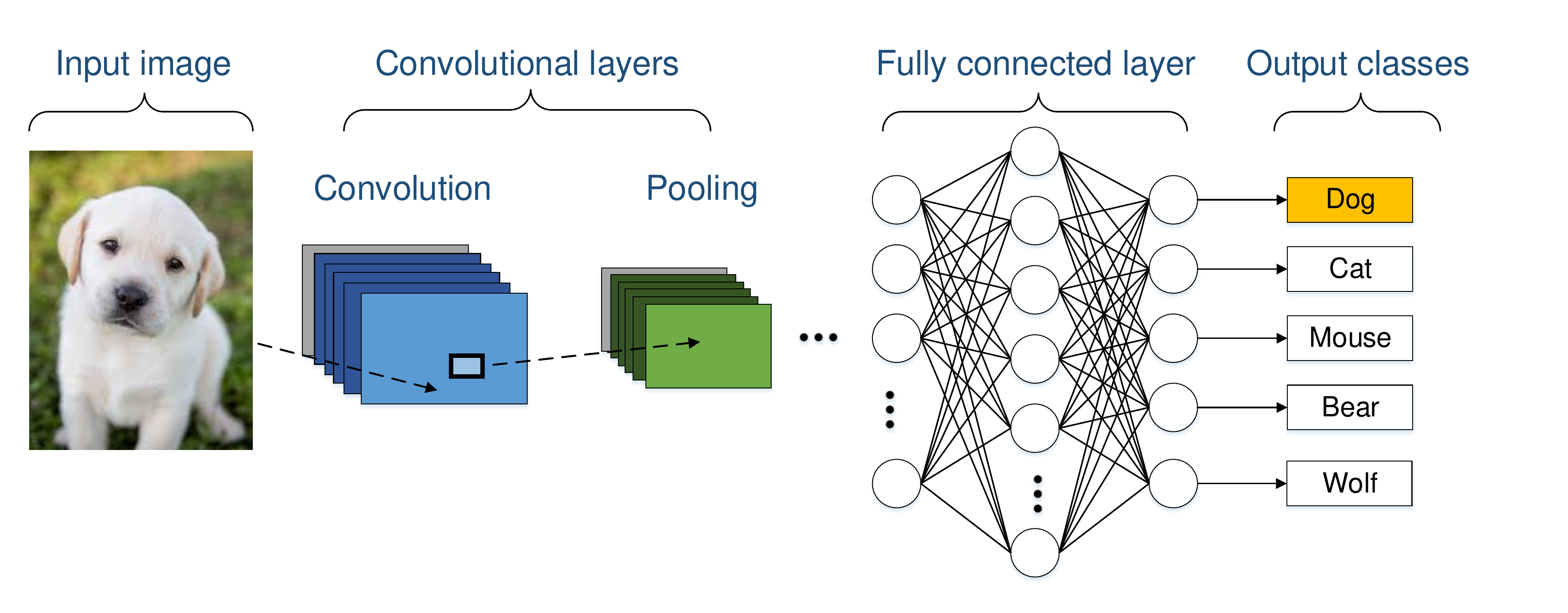}
  \caption{Demonstration of a Convolutional Neural Network (CNN) comprising an input image, convolution and pooling layers, fully connected layer and output classification.}
  \label{fig:cnn}
\end{figure*}

CNNs are categorized as feedforward networks. Fig.~\ref{fig:cnn} shows the most common architecture of a CNN. A CNN consists of one or several convolutional, pooling and fully connected layers. These layers can be connected together to shape a deep convolutional neural network model. The data enters from the input, is processed by various hidden layers and the classification results are presented as outputs.

\subsubsection{\textbf{Convolution Layers}}
The convolutional layers extract features of the input data. A convolutional layer convolves the input with a convolutional kernel (filter) and passes the results to the next layer through a non-linear activation function. More specifically, Each neuron in convolutional layers has a receptive field and connected to other neurons in the adjacent layers through some learned weights and biases. The following equation shows that the feature $F_{k}$ can be computed as:
\begin{equation}
\label{eq:1}
F_{k}  = f \big(W_{k}* I + b_{k} \big) 
\end{equation}
$I$ is the input from the last layer or input image and $W_{k}$ is the convolution kernel for feature $k$ and $b_{k}$ is the bias vector. The Non-linear activation function is denoted as $f(.)$ in \eqref{eq:1}.

\subsubsection{\textbf{Pooling Layers}}
Pooling layers are used in CNNs to down-sample and reduce the spatial resolution of the extracted features. Reduction of spatial resolution can help a CNN model to overcome input distortion and angular transformations. Pooling layers are usually made of max-pooling kernels. A max-pooling kernel selects the maximum value of the region of interest. The max-pooling feature can be defined as:

\begin{equation}
\label{eq:2}
MPF_{k}  = \max_{i,j\in \mathfrak{R_{m,n}}} x_{k, (i,j)}
\end{equation}
where $MPF_{k}$ denotes $k$'th max-pooling feature and $\mathfrak{R_{m,n}}$ shows the region of interest around point $(m,n)$.

\subsubsection{\textbf{Fully Connected Layers}}
After extracting features of the input data by convolutional and pooling layers, the data is sent to a fully connected neural network that implements the decision making process.  The fully connected layer interprets the extracted features and turn them into information presentable with quantities. A softmax function is usually used at the output of the fully connected layer to classify the output.

For more information on CNNs, readers are encouraged to read papers such as \cite{cnn_mit_rev}.

\subsection{\textbf{O\lowercase{pen}CL-based high-level synthesis of CNN\lowercase{s}}}
\subsubsection{\textbf{OpenCL High-level Synthesis on FPGAs}}
OpenCL can be used to write programs that can be executed across heterogeneous platforms. OpenCL offers the ability to describe a parallel algorithm to be implemented on FPGA, for example. 

\begin{figure}[!htb]
  \centering
  \includegraphics[width=0.4\linewidth]{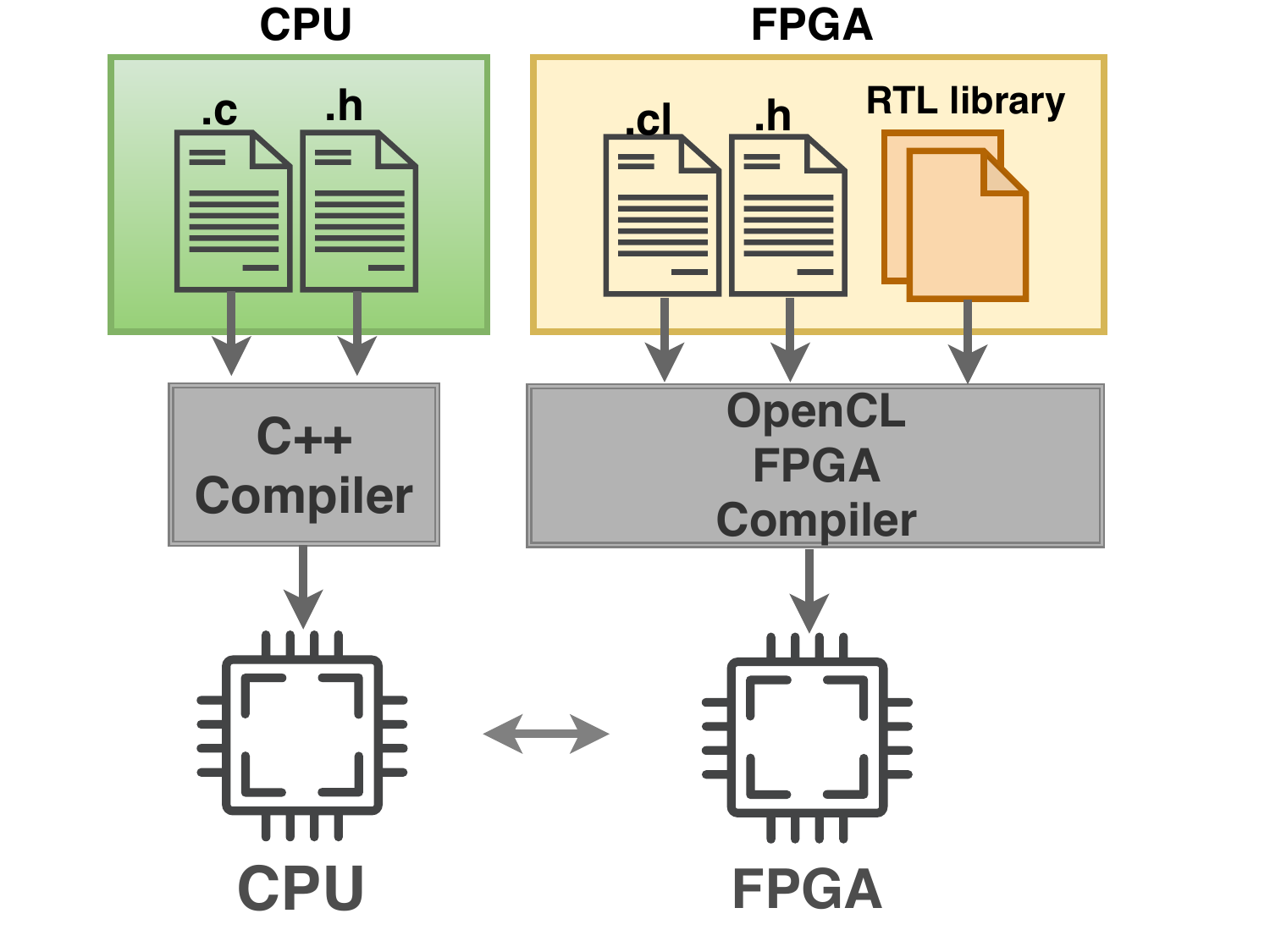}
  \caption{OpenCL high-level synthesis on FPGA.}
  \label{fig:aocl}
\end{figure}

However, this parallel description is at a level of abstraction much higher than hardware description languages such as VHDL. An OpenCL application consists of two parts. The OpenCL ``host'' program that is written purely in C or C++ and that can be executed on any type of processor. On the other hand, ``kernels'' are accelerated functions that are implemented on some co-processor ``device'', such as an FPGA, using fine-grained parallelism. The host can offload computation workload to kernels using sets of command queues. This concept is demonstrated in Fig.~\ref{fig:aocl}.

\subsubsection{\textbf{OpenCL Pipes}} \label{archpipes}

\begin{figure}[!t]
  \centering
  \includegraphics[width=0.5\linewidth]{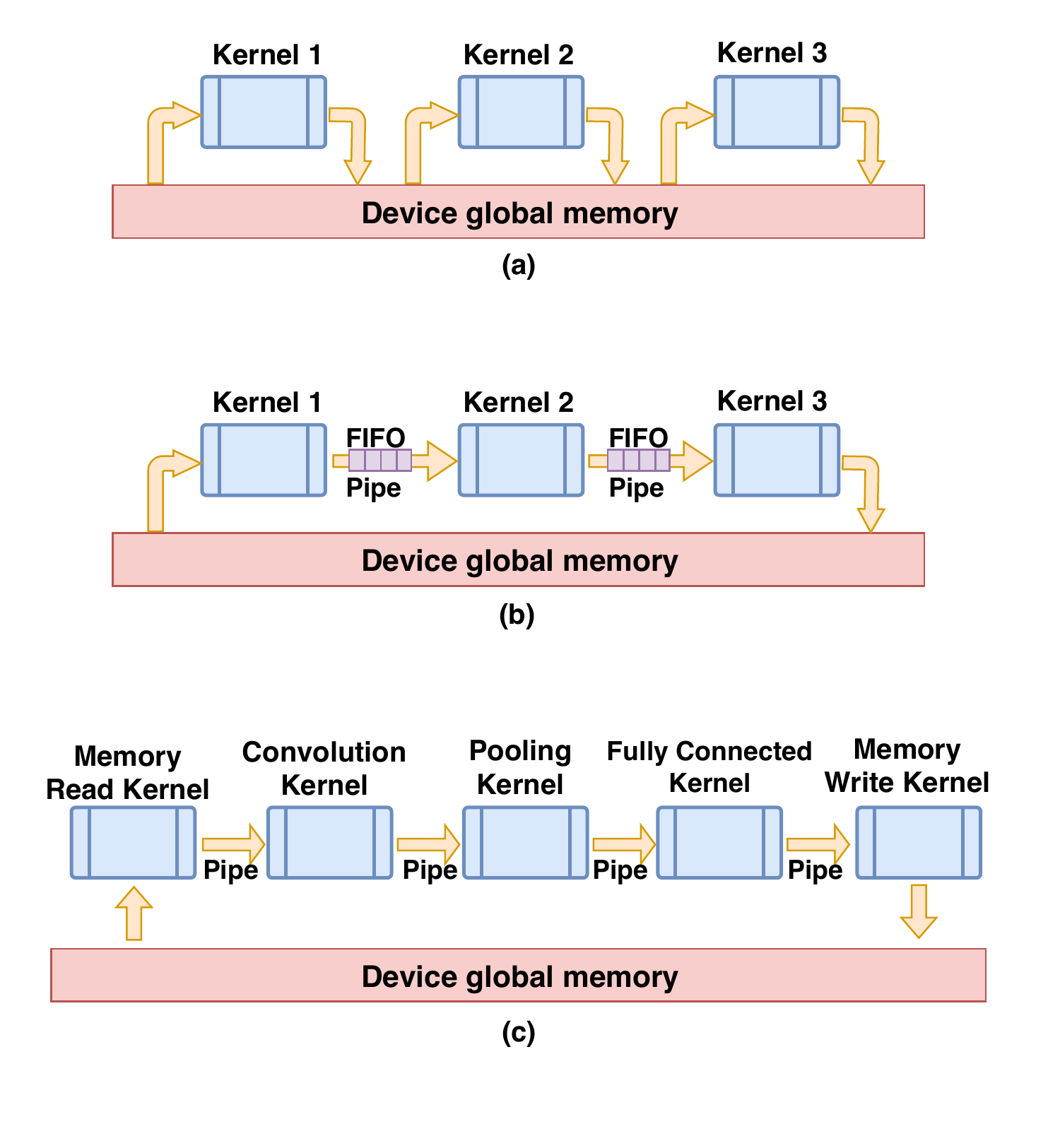}
  \caption{(a) Memory access pattern in OpenCL standard 1.x.
  (b)  OpenCL pipes; in FPGAs, pipes are implemented as FIFOs. 
  (c) Deeply pipelined CNN network architecture.}
  \label{fig:pipes}
\end{figure}

As depicted in Fig.~\ref{fig:pipes}a, in the OpenCL model, every kernel has access to the global memory of the device. In the case of massively parallel algorithms, memory transactions can be a bottleneck. In OpenCL 2.0 \cite{pipes2015opencl}, ``Pipes'' were introduced in standard to enable kernel-to-kernel communications. These tiny communication channels between kernels help to reduce the number of times kernels refer to memory and increases the memory access efficiency. In the case of FPGA, these channels are implemented using FIFOs. Fig.~\ref{fig:pipes}b shows that it is possible to stack OpenCL kernels on top of each other and pass data from one kernel to another. This feature makes FPGAs very well suited to implement stacked layers of a CNN as deeply pipelined kernels. 

\subsubsection{\textbf{OpenCL High-level Synthesis of CNN on FPGA}}

Leveraging the view from Fig.~\ref{fig:pipes}b, a CNN model can be synthesized on FPGA using deep pipelined kernel as shown in Fig.~\ref{fig:pipes}c. For this architecture, the following hardware accelerator kernels are needed: 1) Memory read 2) Memory write 3) Convolution 4) Pooling and 5) Fully connected. In many cases, convolution kernel and the fully connected kernel can be fused together as a single 3-D matrix-matrix multiplication unit. Memory read/write kernels provide and store data for other kernels. This architecture has two main advantages: 1) Depending on the number of computing units in the convolution and pooling layer and the size of data fetched by a memory read/write kernels, this architecture can be scalable (details will be discussed in the design-space algorithm section) and 2) pipelined kernels can process data without storing the data moved between layers; this can significantly improve memory access efficiency.

\begin{figure*}[!htb]
  \centering
  \includegraphics[width=0.35\linewidth]{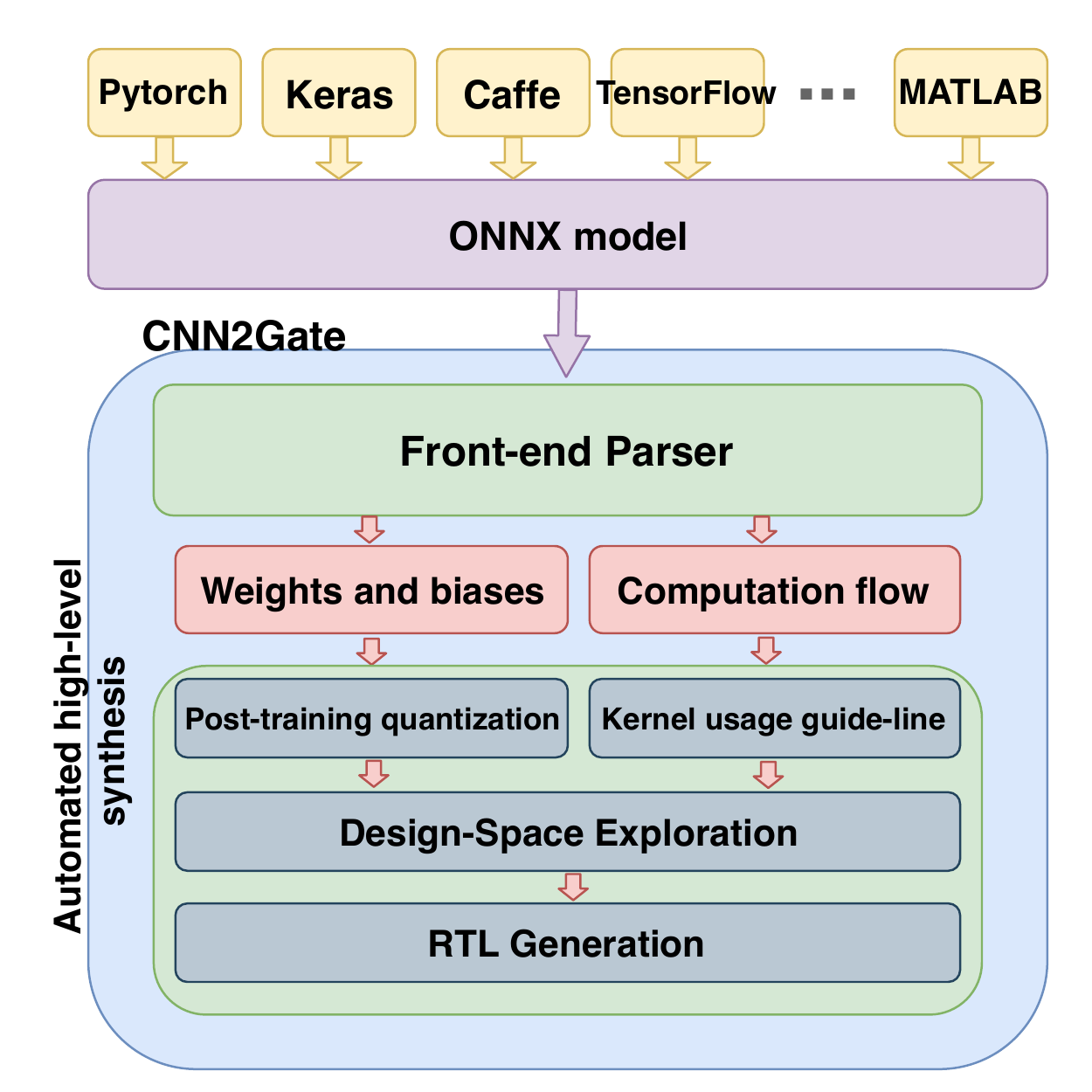} {\textbf{(a)}}
  \includegraphics[width=0.5\linewidth]{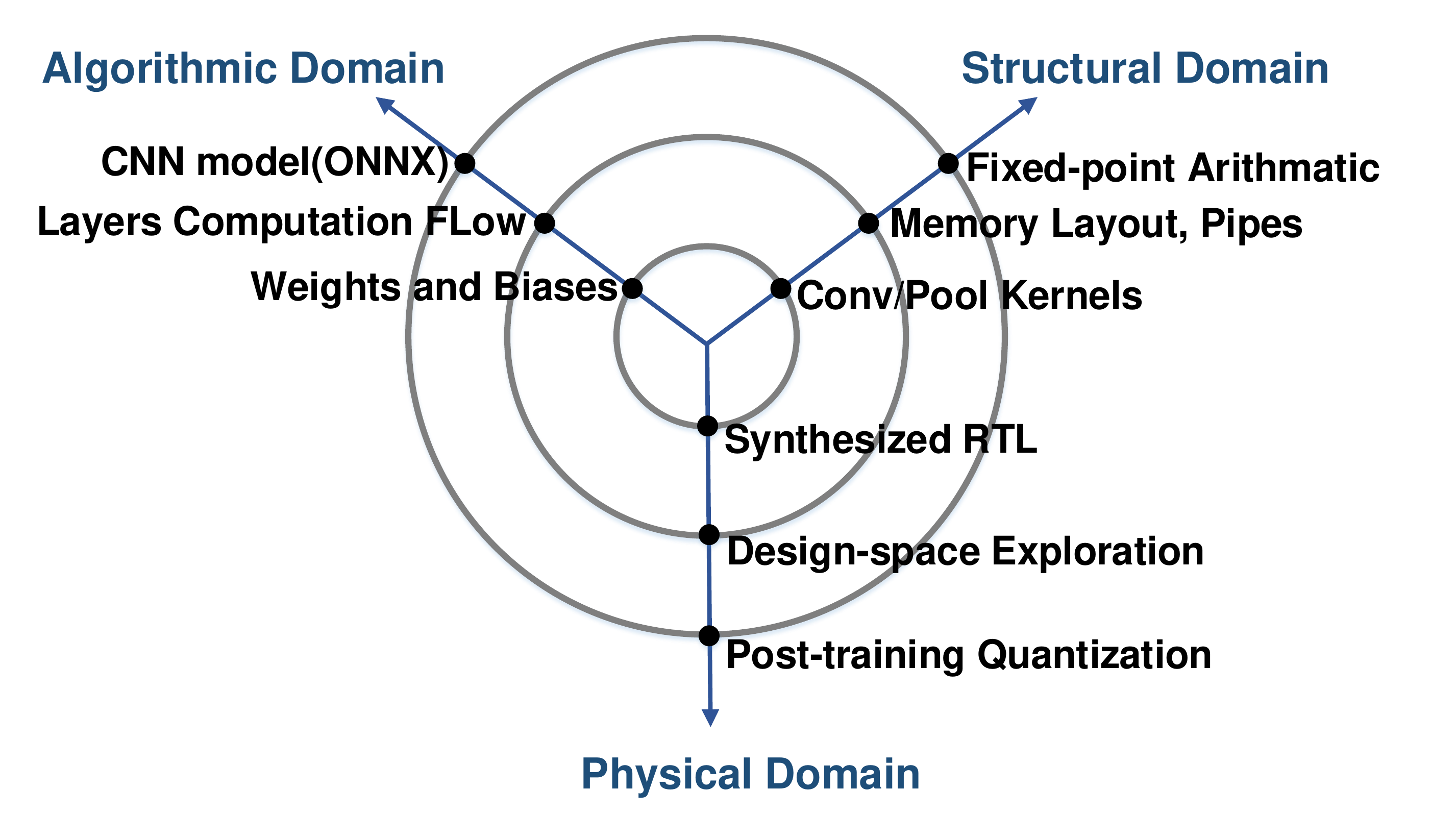} {\textbf{(b)}}
  \vskip 0.1in
  \caption{(a) CNN2Gate overall architecture comprising front-end parser (ONNX parser), design-space exploration, and automated high-level synthesis.
  (b) Gajski-Kuhn chart illustrating different aspects of CNN2Gate.}
  \label{fig:arch}
\end{figure*}

\section{Proposed Architecture} \label{arch}
\subsection{\textbf{Generalized Model Analysis Using ONNX}} \label{onnx}

ONNX is a library that makes deep learning models interoperable. ONNX represents computation flow of a deep neural network as extensible computation graph model using its own built-in operators and data-types. This inter-operable ecosystem eliminates the limitation of using one specific framework among developers. This makes it easier for hardware developers to exploit the proper tool without being bound to the library with which the deep learning model was developed. ONNX provides the definition of a deep neural model using extensible acyclic graphs. Nodes represent an operator and have sets of inputs and outputs. ONNX can support a vast variety of tools such as  Pytorch, TensorFlow, Caffe, Keras and more \cite{onnx}. CNN2Gate offers a front-end ONNX parser for CNNs. As shown in the Fig.~\ref{fig:arch}.a, using ONNX as a transport layer decouples the high-level synthesis tool from the machine learning tool. CNN2Gate parses the computation dataflow --or the arrangement of layers-- besides weights and biases for each layer. The CNN2Gate parser traverses the ONNX graph nodes and extracts the synthesis information of each node based on the following operator types:

\begin{itemize}
    \item \textbf{Convolution}: For the convolution operator, CNN2Gate parses dilations, pads, kernel shape, and stride. The reader can refer to \cite{guide_conv} for more information about these variables and how they affect the computation. It also extracts the learned weights and biases for convolutional kernels. CNN2Gate also computes the output tensor size of the layer using equation \eqref{eq:3}. Let us assume the input of a two dimensional convolutional kernel is of size $(c_{in}, h_{in}, w_{in}) $ where $c$ denotes the number of features, $h$ denotes the height and $w$ denotes the width. The output tensor size $(c_{out}, h_{out}, w_{out}) $ can be written as:
    
\begin{equation}
    \label{eq:3}
    \begin{aligned}
    h_{out} = \left \lfloor \frac{h_{in}+2 p[0]-d[0] (ks[0]-1)-1}{st[0]} +1 \right \rfloor \\
    w_{out} = \left \lfloor \frac{w_{in}+2 p[1]-d[1] (ks[1]-1)-1}{st[1]} +1 \right \rfloor 
    \end{aligned}
\end{equation}

\begin{equation}
\label{eq:4}
c_{out} = c_{in}
\end{equation}

where $ks$ is the kernel size, $st$ is the stride, while $p$ and $d$ are the padding and dilation parameters respectively.
    
    \item \textbf{Max-pooling}: Similar to the convolution, CNN2Gate parses dilations, pads, kernel size, and strides. However, as max-pooling is a down-sampling kernel, it does not have weights and biases. The output tensor size of a max-pooling node with input size of $(c_{in}, h_{in}, w_{in}) $ is identical to equations \eqref{eq:3} and \eqref{eq:4}.
    \item \textbf{ReLu}: CNN2Gate detects the presence of activation function such as ``Relu'' after a convolutional or max-pooling layer.
    \item \textbf{General Matrix Multiplication (\textit{``GEMM''})}: A fully connected layer appears as a GEMM operator in ONNX dataflow graph. In CNN2Gate, there is no specific kernel for the fully connected layer. 
    \item \textbf{Softmax}: CNN2Gate detects the presence of the softmax operator after a fully connected layer.
\end{itemize}

The front-end parser saves layers (nodes) information in a linked structure to preserve the order. Later, this data structure is  used by a high-level hardware synthesis tool. The preserved order serves as a guideline for the synthesizer to configure hardware pipelines.

\subsection{\textbf{Automated High-level Synthesis tool}}

Inspired from the Gajski-Kuhn chart \cite{gajski1983new}, Fig.~\ref{fig:arch}.b sketches the basic idea behind the CNN2Gate automated synthesis workflow. {According to this diagram, the design of CNN2Gate is projected in three domains. The Algorithmic domain axis depicts the definition of concurrent algorithms, the Structural domain axis shows the building blocks to realize the Algorithmic domain. Finally, the Physical domain corresponds to the actual implementations in RTL. The lines connecting the dots show the relations between these domains}. In the ``Algorithmic Domain'', CNN2gate parses the information from an ONNX model as explained in Section \ref{onnx}. In the ``Structural Domain'', CNN2Gate uses 8-bit fixed point arithmetic units to perform computations. In addition, it configures memories, pipes and kernels corresponding to the information that is received from the ONNX model. In the ``Physical Domain'', if weights and biases are float numbers, CNN2Gate can quantize these values based on the information that the user provides from post-training quantization \cite{krishnamoorthi2018quantiziation}. To clarify further, CNN2Gate does not perform quantization itself, however, it can apply a given value that the user provides for a layer. This value can be expressed as an $(N, m)$ pair where fixed-point weights/biases values are represented as $N\times2^{-m}$. Moreover, CNN2Gate performs design-space exploration and generates Register Transfer Level (RTL) models targeting FPGA.

Note that CNN2Gate  can configure several memory buffers depending on the layer operation type automatically. For instance, if the next layer is a fully connected layer, it writes the data to the memory buffer associated with the fully connected layers and similarly, if the layer is convolutional, it writes the data to the convolution buffer.

CNN2Gate is also capable of building and running the CNN model in both emulation and full flow mode. For the emulation mode, CNN2Gate compiles the project for a CPU. It is very important to verify that the CNN model performs correctly in terms of accuracy. The compilation for emulation mode is significantly faster -- in the order of seconds -- as compared to synthesizing the full flow for FPGA, which takes several hours. This feature makes the workflow more versatile for the designers who want to iterate between a FPGA design and a CNN model to reach the best quantization parameters and accuracy. To synthesize the full flow on FPGA, CNN2Gate accepts the name of the FPGA board to perform design-space exploration (Section \ref{fitter}) and generates the RTL accordingly. To validate CNN2Gate, we tested this process by targeting three different Intel\textsuperscript{TM} FPGA boards \cite{de0nano,de1soc,nallatech510} and report the results later in this paper. In addition, we used Intel OpenCL SDK 16.1 to synthesize our designs.  This workflow can be reproduced for Xilinx\textsuperscript{TM} FPGAs with a little effort.

\begin{figure*}[!htb]
  \centering
    \begin{minipage}{\textwidth}
  \includegraphics[width=0.95\linewidth]{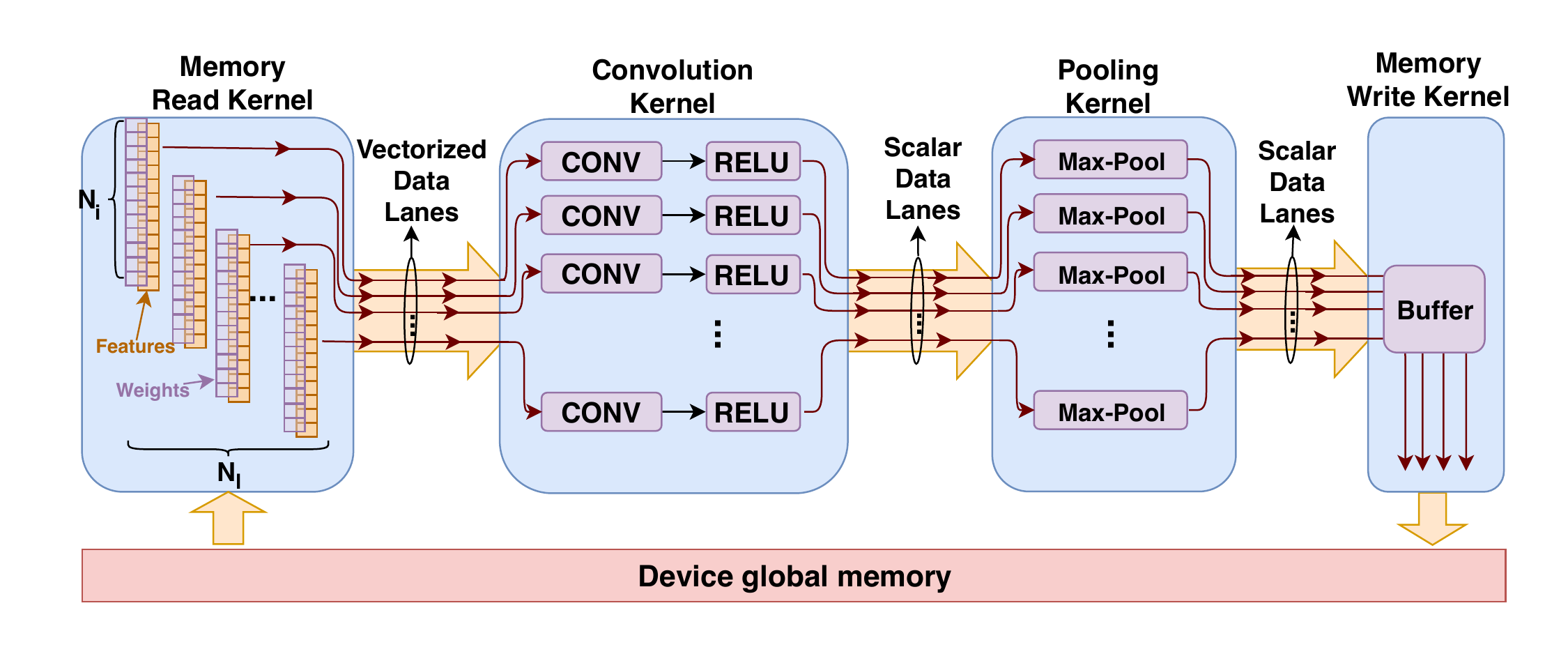} 
   \caption{Detailed demonstration of vectorized input data and weights and the concept of computation lanes in pipelined kernels.}
  \label{fig:fitter}
  \end{minipage}
\end{figure*}

Taking advantage of  OpenCL flexibility, it is possible to design an architecture with several degrees of freedom. The first degree of freedom is the size of the pipes. The better throughput of the pipes means less congestion point for data to be moved from a kernel to another. The second degree of freedom is the bandwidth of data that memory write/read kernels provide. The third degree of freedom is the number of parallel {convolutional (CONV) and RELU} units that are used to implement convolution kernels. Fig.~\ref{fig:fitter} shows this concept in a simple example. These degrees of freedom for deeply pipelined kernels are leveraged from what proposed by \cite{2017pipecnn}. The memory read kernel fetches $N_{l}$ vectors of size $N_{i}$  for features and weights. Tuning $N_{i}$ and $N_{l}$ can provide a better throughput for data write/read kernels. {Note that the memory access schedule of where and when to read the features and weights are derived  by the host program. The memory access schedule is configured by the front-end parser based on the CNN model }. 
The number of computation lanes ($N_{l}$) shows the level of parallelism of the algorithm. The number of CONVs in a convolution kernel, the size of data pipes and the number of max-pool operators in the max-pool kernel are tuned according to $N_{l}$. Changing $N_{l}$ and $N_{i}$ can  results in different utilization ratios of FPGA resources. For instance, in the case of increasing  $N_{l}$, 1) more on-chip memory in read/write kernels, 2) more register for pipe FIFOs, 3) more DSP slices for CONVs and 4) more LUT for max-pooling kernels are needed to accommodate the design on FPGA. It is worth mentioning that arbitrary choices for  $N_{l}$ and  $N_{i}$ are not always possible.  $N_{i}$ should be a divisor of the features' width for all layers to avoid padding. Likewise,  $N_{l}$ should be a divisor of the number of features for all layers to avoid idle lanes in some layers. 

\subsection{\textbf{Hardware-aware Design-space Exploration}} \label{fitter}
CNN2Gate analyses the  computation flow layers that are extracted from the model by the ONNX front-end parser and determines all options possible for $N_{l}$ and  $N_{i}$. CNN2Gate is also capable of interacting with the Intel OpenCL compiler to estimate resource usage for a specific choice of $N_{l}$ and  $N_{i}$.
Considering the dynamics of our problem, we suggest two methods for design-space exploration.

\subsubsection{\textbf{Brute Force Design Space Exploration (BF-DSE)}}
    
    This method exhaustively searches for all possible pairs of  $N_{l}$ and  $N_{i}$ and finds the feasible option that maximizes FPGA resource utilization. In our case, the solution maximizing resource utilization corresponds to the one providing the best throughput. This method is simple to execute and it always find the best solutions. However, for larger FPGAs with large number of a possible candidates for  $N_{l}$ and  $N_{i}$ it can take longer time to perform the search. In the next section, a more efficient search strategy is described.

\subsection{\textbf{Reinforcement Learning Design Space Exploration (RL-DSE)}}
    
RL-DSE trains a reinforcement learning (RL) agent to find the best options for $N_{l}$ and  $N_{i}$. Exploiting reinforcement learning is of interest design-space exploration for two reasons. First, it can be faster than brute force and second, it can be merged with other RL-agents such as HAQ \cite{haq} or ReleQ \cite{releq} to determine the level of parallelism and the quantization of each layer. RL-DSE explores hardware options for $N_{l}$ and  $N_{i}$ and finds the best fit for the specified hardware. The RL-DSE agent receives a reward signal corresponding to the feed-back provided by the Intel OpenCL compiler for FPGA resource utilization. Hence, the reward function for RL-DSE is designed to maximize FPGA resource utilization.

When CNN2Gate contacts Intel OpenCL compiler to evaluate a hardware option, it receives the corresponding hardware resource utilization. This feedback comprises 1) the percentage of look up table utilization, 2) the percentage of DSP utilization, 3) the percentage of on-chip memory block utilization and 4) the percentage of register utilization. We denote these percentages as $P_{lut}$, $P_{dsp}$, $P_{mem}$ and $P_{reg}$ respectively.

Given the percentages of the utilization, the agent takes a series of actions. The agent starts from the minimum values of $N_{l}$ and  $N_{i}$. RL-DSE can flexibly choose to 1) increase $N_{l}$, 2) increase $N_{i}$, or 3) increase both $N_{i}$ and $N_{l}$. If one of the variables reaches the maximum possible value based on the CNN topology, the variable is  reset to its initial value.

Let us assume that the average usage factor is defined as:
\begin{equation}
\label{eq:5}
F_{avg} = \frac{P_{lut} + P_{dsp} + P_{mem} +P_{reg}}{4}
\end{equation}
Further assume that the user defines {a vector of thresholds} $T_{th}$ for the maximum usage that is tolerated for each quota. The reward function can be described as follows.

{
\hfill{
\begin{minipage}{.53\linewidth}
\centering
\begin{algorithm}[H]
    \SetKwInOut{Input}{Input}
    \SetKwInOut{Output}{Output}

    \Input{$T_{th}$ , $F_{avg}$, $P_{lut}$, $P_{dsp}$, $P_{mem}$, $P_{reg}$, $N_{i}$ and $N_{l}$}
    \Output{$Reward$ , $H_{best}$}
    \textbf{Variables: $F_{max}, T_{th} = (T_{lut},T_{dsp},T_{mem},T_{reg})$}\\
    \eIf{{($P_{lut}$, $P_{dsp}$, $P_{mem}$, $P_{reg}$)} < $(T_{lut},T_{dsp},T_{mem},T_{reg})$}
      {
        \eIf{$F_{avg}$ > $F_{max}$}
          {
            $F_{max}$ = $F_{avg}$\\
            $Reward = \beta \times F_{avg}$\\
            $H_{best} = (N_{i}, N_{l})$
          }
          {
            $Reward = 0$
          }
      }
      {
        $Reward = -1$
      }
    \caption{Reward shaping}
    \label{alg:reward}
\end{algorithm}
\end{minipage}
}
}\hfill \break

In  Algorithm \ref{alg:reward}, $H_{best}$ is the best hardware options and $F_{max}$ is the maximum usage degree observed by the agent during the exploration of the search environment. In the reward shaping function, {if at least one of the hardware utilization quotas ($P_{lut}$, $P_{dsp}$, $P_{mem}$, $P_{reg}$) exceeds the thresholds specified in the vector of threshold limits ($T_{th} = (T_{lut},T_{dsp},T_{mem},T_{reg})$),} the agent receives a negative reward for this choice and the chance of choosing this option for a later iteration decreases. The reward function keeps track of maximum usage score $F_{max}$, and constantly update $F_{max}$ and $H_{best}$ with the best hardware option observed by the agent in the environment. Since during the exploration, the best value of the reward function is unknown, there is no actual stop condition for agent exploration. In this case, we used a variation of time-limited reinforcement learning \cite{mnih2016asynchronous} with which, the number of iterations in each episode is limited. Finally, in our RL-DSE,  {a scaling factor $\beta=0.01$ is applied to $F_{avg}$ to form the final reward function to convert it from percentage scale to a number between 0 and 1.}

Note that a discount factor $\gamma = 0.1$ is used in our RL agent design. {The discount factor specifies that the agent does not have unlimited time to find the best option. Thus, the agent receives a down-scaled reward as it spends more time in the environment. The discount factor urges the agent to optimize the total discounted reward} \cite{van2007reinforcement}:
\begin{equation}
\label{eq:6}
r_{t} + \gamma r_{t+1} + \gamma^2 r_{t+1} + ... = \sum_{i=1}^n \gamma^i r_{t+i}
\end{equation}
{where $r_{t}$ is the reward calculated in time $t$ according to Algorithm} ~\ref{alg:reward}.

\section{Results}\label{res}
Table \ref{ExecTimeTable} shows the execution times of  AlexNet \cite{alexnet} and VGG-16 \cite{vgg} for three platforms using CNN2Gate. As mentioned before, CNN2Gate is capable of verifying the design in emulation mode with a  CPU. Even if the execution time is rather large, this is a very useful feature to let the developer verify the validity of the CNN design on the target hardware before going forward for synthesis, which is a very time-consuming process. Note that the emulation's execution time cannot be a reference for the throughput performance of a core-i7 processor. The emulation mode only serves the purpose of verifying the OpenCL kernels operations. In \cite{cpu_benchmarking}, the authors described the execution of AlexNet on desktop CPUs and reported an execution time as low as 2.15 seconds. The reported results also show the scalability of this design Indeed, results are reported for both the low cost Cyclone V SoC and for the much more expensive Arria 10 FPGA that has much more resources. The exploited level of parallelism was automatically extended by the design-space exploration algorithm to obtain better results for execution times commensurate with the capacity of the hardware platform.
\begin{table}[H]
\renewcommand\thetable{1}
\centering
\caption{Execution times for Alexnet and VGG \small{({batch size = 1})}}
\label{ExecTimeTable}
\setlength{\tabcolsep}{3pt}
\begin{tabular}{|>{\centering}m{50pt}|>{\centering}m{60pt}|>{\centering}m{35pt}| >{\centering}m{35pt}|>{\centering\arraybackslash}m{20pt}|}
\hline
\textbf{Platform} & \textbf{Resource Utilization$^1$} & \multicolumn{2}{c|}{\textbf{Execution time}}      & $f_\mathit{max}$ \\
  && AlexNet & VGG-16     &  \\
\hline
\hline
\textbf{Core-i7} \par (Emulation)& 
N/A & 13 s & {148 s} & N/A \\
\hline
\textbf{Cyclone V SoC} \par {5CSEMA5} & 
\textbf{Logic:} 83 \% \par \textbf{DSP:}  83 \% \par \textbf{RAM blocks:} 100 \% & 153 ms & {4.26 s} & 131 MHz \\
\hline
\textbf{Arria 10} \par {GX 1150} & 
\textbf{Logic:} 30\% \par \textbf{DSP:}  20 \% \par \textbf{RAM blocks:} 40 \%& 18 ms & {205 ms} & 199 MHz \\
\hline
\end{tabular}

$^1$for AlexNet

\end{table}

\begin{table*}[!b]
\renewcommand\thetable{2}
\centering
\caption{CNN2Gate Synthesis and Design-Space Exploration Details {(AlexNet)}}
\label{DSEtable}
\setlength{\tabcolsep}{3pt}
\centering
\begin{tabular}{|>{\centering}m{50pt}|>{\centering}m{40pt}|>{\centering}m{40pt}| >{\centering}m{40pt}|>{\centering}m{80pt}| >{\centering}m{80pt}| >{\centering\arraybackslash}m{50pt}|}
\hline
\textbf{Platform} & \textbf{RL-DSE time} & \textbf{BF-DSE time}   & \textbf{Synthesis time} & \textbf{Resources Available} & \textbf{Resources Consumed} & \textbf{Hardware Options} ($N_i$, $N_l$)\\
\hline
\hline
\textbf{Cyclone V SoC} \par {5CSEMA4} & 
2.5 min& 3.5 min & N/A & \textbf{ALM:} 15 K \par \textbf{DSP:}  83  \par \textbf{RAM blocks:} 321&  Does not fit & N/A \\
\hline
\textbf{Cyclone V SoC} \par {5CSEMA5} & 
2.5 min& 3.5 min & 46 min & \textbf{ALM:} 32 K \par \textbf{DSP:}  87 \par \textbf{RAM blocks:} 397 \par \textbf{Mem. bits:} 4 M& 
\textbf{ALM:} 26 K \par \textbf{DSP:}  72 \par \textbf{RAM blocks:} 397 \par \textbf{Mem. bits:} 2 M& (8,8) \\
\hline
\textbf{Arria 10} \par {GX 1150} & 
3 min & 4 min & 8.5 hrs & \textbf{ALM:} 427 K \par \textbf{DSP:}  1516 \par \textbf{RAM blocks:} 2713 \par \textbf{Mem. bits:} 55.5 M &  
\textbf{ALM:} 129 K \par \textbf{DSP:} 300 \par \textbf{RAM blocks:} 1091\par \textbf{Mem. bits:} 16 M & (16,32) \\
\hline
\end{tabular}
\end{table*}

Note that resource consumption of the Arria 10 FPGA remained fairly low (less than 40 percent). This is because the design-space exploration algorithm of the implemented prototype has limited options to attempt using the hardware platform to its full extent when implementing a given design as explained in  Section \ref{fitter}. To maintain the scalability of the algorithm, it is not always possible to use arbitrary choices for $(N_i , N_l)$ parameters. {These parameters must be chosen in a way that kernels can be used  as the building block of all the layers}. This leads to have limited options to increase the level of parallelism that can be exploited with a given network algorithm. Relaxing this limitation, {(i.e. manually designing kernels based on each layers' computation flow)}  could lead to higher resources consumption and higher throughput at the expense of losing {the scalability that is needed for automation of this process}. The maximum operating frequency ($f_{max}$)  varies for different FPGAs. Indeed it depends on the underlying technology and FPGA family. Intel OpenCL compiler (synthesizer) automatically adjust PLLs on the board to use the maximum frequency for the kernels. It is of interest that the operating frequency of the kernels supporting AlexNet and VGG-16 were the same as they had essentially the same critical path, even though VGG-16 is a lot more complex. The larger complexity of VGG was handled by synthesizing a core (i.e. Fig.~\ref{fig:fitter} ) that executes a greater number of cycles if the number of layers in the network increases.
\begin{figure*}[!t]
  \centering
\begin{minipage}{0.9\textwidth}
  \includegraphics[width=1\linewidth]{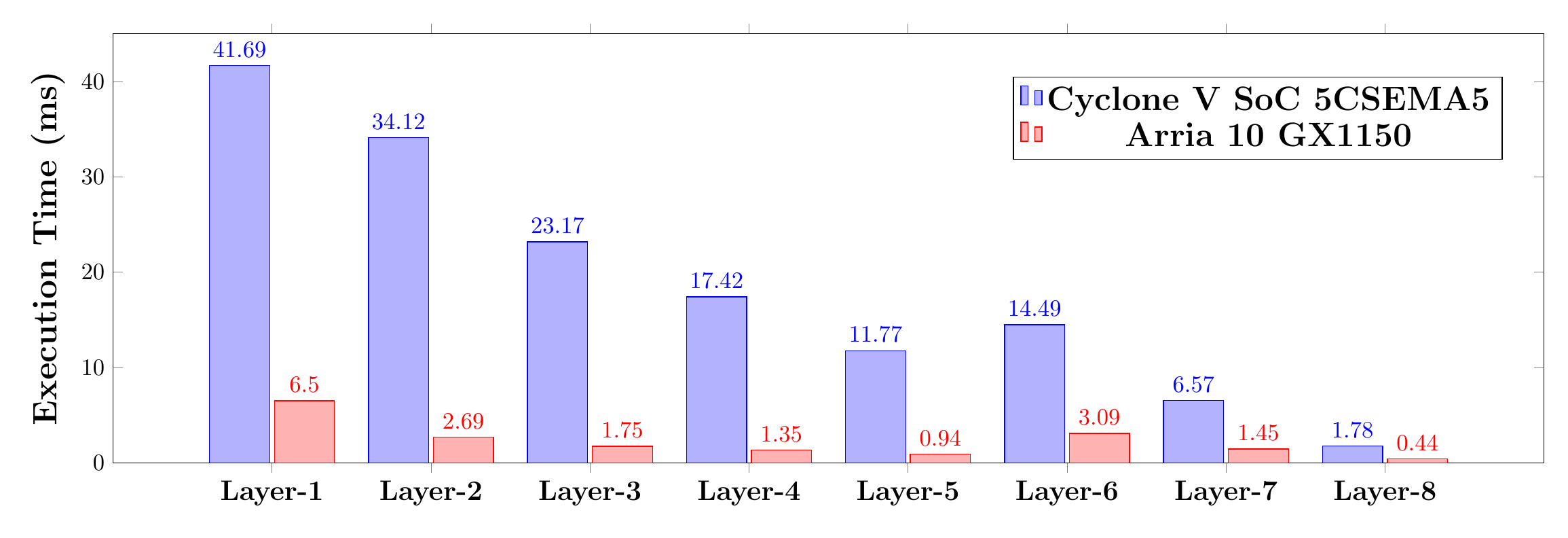} 
   \caption{Detailed break-down of execution time for each layer of AlexNet. A Layer here means execution of one round of pipelined kernels as shown in Fig.~\ref{fig:fitter}. In case of AlexNet Layer-1 to Layer-5 are combination of memory read/write, convolution and pooling kernels while Layer-5 to Layer-7 are combination of memory read/write and a fully connected kernels.}
  \label{fig:bar_execution}
  \end{minipage}
\end{figure*}

\begin{table*}[b]
\renewcommand\thetable{3}
\centering
\caption{Comparison to the existing works {AlexNet} with hardware options $(N_i, N_l)=(16,32)$}
\label{ExistingWorkTable}
\setlength{\tabcolsep}{3pt}
\centering
\begin{tabular}{|>{\centering}m{90pt}||>{\centering}m{50pt}|>{\centering}m{50pt}| >{\centering}m{50pt}|>{\centering}m{50pt}|>{\centering\arraybackslash}m{80pt}|}
\hline
 & \small{AlexNet}\cite{zhang2015optimizing} & \small{AlexNet}\cite{ma2016scalable}   & \small{AlexNet}\cite{fpgaconvnet_1_2018} & \small{AlexNet}\cite{suda2016throughput} & AlexNet \par [This work]*\\
\hline
\hline
\textbf{FPGA}&{Virtex-7 VX485T}&Stratix-V GXA7&Zynq \par 7045&Stratix-V GX-D8&Arria 10 GX1150\\
\hline
\textbf{Synthesis method}&C/C++&RTL&C/C++&OpenCL&OpenCL\\
\hline
\textbf{Frequeny} \small{(MHz)}&100&100&125&-&199\\
\hline
\textbf{Logic Utilization}&186K (61\%) &121K (52\%) & -& 120K (17\%) &129K (30\%)\\
\hline
\textbf{DSP Utilization}&2,240 (80\%)&256 (100\%) &897(99.5\%)& 665 (34\%)&300 (20\%)\\
\hline
\textbf{Latency} \small{(ms)}&21.61&12.75&8.22&20.1& 18.24\\
\hline
\textbf{Precision} \small{(bits)} &32 float&8-16 fixed& 16 fixed&8-16 fixed& 8 fixed\\
\hline
\textbf{Performance} \small{(GOp/s)}&61.62&114.5&161.98&72.4&80.04\\ 
\hline
\end{tabular}

*batch size = 1
\end{table*}

\begin{table*}[t]
\renewcommand\thetable{4}
\centering
\caption{Comparison to the existing works {VGG-16}  with hardware options $(N_i, N_l)=(16,32)$}
\label{ExistingWorkTableVGG}
\setlength{\tabcolsep}{3pt}
\centering
\begin{tabular}{|>{\centering}m{90pt}||>{\centering}m{50pt}|>{\centering}m{50pt}| >{\centering}m{50pt}|>{\centering}m{50pt}|>{\centering\arraybackslash}m{80pt}|}
\hline
 & \small{VGG-16}\cite{qiu2016going} & \small{VGG-16}\cite{ma2017optimizing}   & \small{VGG-16}\cite{fpgaconvnet_1_2018} & \small{VGG-16}\cite{suda2016throughput} & VGG-16 \par [This work]*\\
\hline
\hline
\textbf{FPGA}&{Zynq \par7045}&Arria 10 GX1150&Zynq \par 7045&Stratix-V GX-D8&Arria 10 GX1150\\
\hline
\textbf{Synthesis method}&-&RTL&C/C++&OpenCL&OpenCL\\
\hline
\textbf{Frequeny} \small{(MHz)}&150&150&125&120&199\\
\hline
\textbf{Logic Utilization}&182 (83.5\%) &161K(38\%) &-& - &129K (30\%)\\
\hline
\textbf{DSP Utilization}&780 (89.2\%)& 1518(100\%) &855(95\%)& -&300 (20\%)\\
\hline
\textbf{Latency} \small{(ms)}&-&47.97&249.5&262.9& 205 \\
\hline
\textbf{Precision} \small{(bits)} &16 fixed&8-16 fixed& 16 fixed&8-16 fixed& 8 fixed\\
\hline
\textbf{Performance} \small{(GOp/s)}&136.91&645.25&161.98&117.8&151.7\\ 
\hline
\end{tabular}

*batch size = 1
\end{table*}

Table \ref{DSEtable} gives more details regarding the design-space exploration algorithms which are coupled with synthesis tool. Both RL-DSE and BF-DSE algorithm use the resource utilization \textit{estimation} of the synthesizer to fit the design on the FPGA. This is important as the time consumed for design-space exploration is normally under 5 minutes, while the synthesis time for larger FPGAs such as the Arria 10, can be close to 10 hours. Analyzing brute-force and reinforcement learning executing times shows that brute-force algorithm is almost 25 percent slower than the reinforcement learning algorithm. This time is negligible compared to the synthesis time that is in order of hours.{ The goal of suggesting the RL-DSE method is to demonstrate that it is possible to further decrease the exploration time by using better search methods}. The RL-DSE algorithm would be more valuable if it could be exploited in conjunction to the reinforcement learning quantization algorithms such as ReLeQ \cite{releq}. However, it is good to have the time of design space exploration reduced as the user may use this algorithm several times in {the design stage for various CNN models before synthesis.}
We tried CNN2Gate on three platforms. The first one is a relatively small Cyclone V device with 15K adaptive logic modules (ALMs) and 83 DSPs. The fitter could not fit either ALexNet or VGG on this device. Clearly, the minimum space required for fitting this design on FPGA due to control logic is fairly large. CNN2Gate did not experience any difficulty fitting the design on bigger FPGAs as demonstrated in the  {Table}~\ref{DSEtable}. Resource utilization and hardware options $(N_i , N_l)$ are also provided, which correspond to the execution times shown in Table~\ref{ExecTimeTable}.

CNN2Gate resource consumption is very similar for AlexNet and VGG-16. In case of identical hardware options, CNN2Gate's synthesized core is going to be nearly identical for all CNN architecture as shown in Fig.~\ref{fig:fitter}. The only difference is the size of internal buffers to allocate the data in the computation flow. More quantitatively, in our implementation, VGG-16 uses 8\% more of the Arria 10 FPGA block RAMs {in comparison to what is shown in Table}~\ref{DSEtable} {for AlexNet}.

Revisiting Fig.~\ref{fig:fitter}, pipelined kernels are capable of reading data from global memory and process the convolution and pooling kernel at once. In addition, for fully connected layers in CNNs, the convolution kernel acts as the main data process unit and the pooling kernel is configured as a pass-through. Considering this hardware configuration, we can merge convolution and pooling layers as one layer. In the case of AlexNet, this leads to five fused convolution/pooling and three fully-connected layers. Fig.~\ref{fig:bar_execution} reports the detailed execution time of this six layers including memory read and write kernels for each layer. Thus, as the algorithm proceed through the layers, dimensions of the data (features) reduced and the execution time is decreased.

Table \ref{ExistingWorkTable} shows a detailed comparison of CNN2Gate to other existing works for {AlexNet}. CNN2Gate is faster than \cite{zhang2015optimizing, suda2016throughput} in terms of latency and throughput. However, CNN2Gate uses more FPGA resources than \cite{suda2016throughput}. To make a fair comparison, we can measure relative performance density per DSP or ALMs. In this case, CNN2Gate performance density (GOp/s/DSP) is higher (0.266) when compared to 0.234 for \cite{suda2016throughput}. There are other designs such as \cite{ma2016scalable, fpgaconvnet_1_2018} which are faster than our design in terms of pure performance.Nevertheless, the CNN2Gate method outlined above meaningfully improves on existing methods as it more scalable and automatable than the methods presented in \cite{ma2016scalable, fpgaconvnet_1_2018}, which are limited in these regards as they require human intervention in order to reach high levels of performance.
Second, the design entry of \cite{ma2016scalable, fpgaconvnet_1_2018} are RTL and C respectively, while our designs were automatically synthesized from ONNX using OpenCL. Thus, not surprisingly, our work does not achieve the maximum reported performance. This is partly due to the use of ONNX as a starting point, and trying to keep the algorithm scalable for either large and small FPGAs. This can impose some limitations such as the maximum number of utilized CONV units per layer. There are also other latency reports in the literature such as \cite{2017pipecnn}. However, those latency reports are measured in the favorable batch size (e.g. 16). Increasing batch size can make more parallelism available to the algorithm that can lead to higher throughput. {However, for clarity, we limited the comparisons in Table}~ \ref{ExistingWorkTable} and 4 {to batch size = 1}.

Table \ref{ExistingWorkTableVGG} shows a detailed comparison of CNN2Gate to other existing works for VGG-16. It is observable that CNN2Gate is performing better for larger neural networks such as VGG. CNN2Gate achieves 18 \% lower latency than \cite{fpgaconvnet_1_2018}, despite the fact that CNN2Gate is using fewer DSPs. While for AlexNet, \cite{fpgaconvnet_1_2018} was more than 50 \% faster than CNN2Gate. Finally, for VGG-16, we did find some hand tailored RTL custom designs such as \cite{ma2017optimizing} that are faster than CNN2Gate.

\section{Conclusion}
This paper describes the design and implementation of a general framework for implementing convolutional neural networks on FPGAs. This framework takes the form of a Python library that can be integrated into a wide range of popular machine learning frameworks. CNN2Gate makes it easy for machine learning developers to program and use FPGAs to perform inference. CNN2Gate exploits the OpenCL synthesis workflow for FPGAs offered by commercial vendors. CNN2Gate is capable of parsing the data-flow of CNN models expressed in ONNX. This framework also has an integrated design-space exploration tool that helps developers finding the best hardware option for synthesizing a given CNN model on an FPGA. CNN2Gate achieves a classification latency of 205ms for VGG-16 and 18ms for AlexNet on Intel Arria 10 FPGA. These results are excellent considering they are obtained by an automated design space exploration and synthesis process which does not rely on expert's low-level hardware knowledge.

\bibliographystyle{unsrt}  
\bibliography{CNN2Gate}  

\begin{thebibliography}{10}

\bibitem{cnn_mit_rev}
Waseem Rawat and Zenghui Wang.
\newblock Deep convolutional neural networks for image classification: A
  comprehensive review.
\newblock {\em Neural computation}, 29(9):2352--2449, 2017.

\bibitem{strigl2010performance}
Daniel Strigl, Klaus Kofler, and Stefan Podlipnig.
\newblock Performance and scalability of gpu-based convolutional neural
  networks.
\newblock In {\em 2010 18th Euromicro Conference on Parallel, Distributed and
  Network-based Processing}, pages 317--324. IEEE, 2010.

\bibitem{krishnamoorthi2018quantiziation}
Raghuraman Krishnamoorthi.
\newblock Quantizing deep convolutional networks for efficient inference: A
  whitepaper.
\newblock {\em arXiv preprint arXiv:1806.08342}, 2018.

\bibitem{8bitq}
Naigang Wang, Jungwook Choi, Daniel Brand, Chia-Yu Chen, and Kailash
  Gopalakrishnan.
\newblock Training deep neural networks with 8-bit floating point numbers.
\newblock In {\em Advances in neural information processing systems}, pages
  7675--7684, 2018.

\bibitem{intel2017can}
Eriko Nurvitadhi, Ganesh Venkatesh, Jaewoong Sim, Debbie Marr, Randy Huang,
  Jason Ong Gee~Hock, Yeong~Tat Liew, Krishnan Srivatsan, Duncan Moss, Suchit
  Subhaschandra, et~al.
\newblock Can fpgas beat gpus in accelerating next-generation deep neural
  networks?
\newblock In {\em Proceedings of the 2017 ACM/SIGDA International Symposium on
  Field-Programmable Gate Arrays}, pages 5--14. ACM, 2017.

\bibitem{intelsoc}
Intel.
\newblock {Intel User-Customizable Soc FPGAs}.
\newblock
  \url{https://www.intel.com/content/dam/www/programmable/us/en/pdfs/literature/wp/wp-01167-custom-arm-soc.pdf},
  2019.

\bibitem{2017pipecnn}
Dong Wang, Ke~Xu, and Diankun Jiang.
\newblock Pipecnn: An opencl-based open-source fpga accelerator for convolution
  neural networks.
\newblock In {\em 2017 International Conference on Field Programmable
  Technology (ICFPT)}, pages 279--282. IEEE, 2017.

\bibitem{fpgaconvnet_1_2018}
S.~I. Venieris and C.~S. Bouganis.
\newblock {fpgaConvNet: Mapping Regular and Irregular Convolutional Neural
  Networks on FPGAs}.
\newblock {\em IEEE Transactions on Neural Networks and Learning Systems},
  pages 1--17, 2018.

\bibitem{umuroglu2017finn}
Yaman Umuroglu, Nicholas~J Fraser, Giulio Gambardella, Michaela Blott, Philip
  Leong, Magnus Jahre, and Kees Vissers.
\newblock Finn: A framework for fast, scalable binarized neural network
  inference.
\newblock In {\em Proceedings of the 2017 ACM/SIGDA International Symposium on
  Field-Programmable Gate Arrays}, pages 65--74. ACM, 2017.

\bibitem{ma2017optimizing}
Yufei Ma, Yu~Cao, Sarma Vrudhula, and Jae-sun Seo.
\newblock Optimizing loop operation and dataflow in fpga acceleration of deep
  convolutional neural networks.
\newblock In {\em Proceedings of the 2017 ACM/SIGDA International Symposium on
  Field-Programmable Gate Arrays}, pages 45--54. ACM, 2017.

\bibitem{bilaniuk2019bit}
Olexa Bilaniuk, Sean Wagner, Yvon Savaria, and Jean-Pierre David.
\newblock Bit-slicing fpga accelerator for quantized neural networks.
\newblock In {\em 2019 IEEE International Symposium on Circuits and Systems
  (ISCAS)}, pages 1--5. IEEE, 2019.

\bibitem{pipes2015opencl}
Jasmina Vasiljevic, Ralph Wittig, Paul Schumacher, Jeff Fifield,
  Fernando~Martinez Vallina, Henry Styles, and Paul Chow.
\newblock Opencl library of stream memory components targeting fpgas.
\newblock In {\em 2015 international conference on field programmable
  technology (FPT)}, pages 104--111. IEEE, 2015.

\bibitem{hls4ml}
Javier Duarte, Song Han, Philip Harris, Sergo Jindariani, Edward Kreinar,
  Benjamin Kreis, Jennifer Ngadiuba, Maurizio Pierini, Ryan Rivera, Nhan Tran,
  et~al.
\newblock Fast inference of deep neural networks in fpgas for particle physics.
\newblock {\em Journal of Instrumentation}, 13(07):P07027, 2018.

\bibitem{zhang2018caffeine}
Chen Zhang, Guangyu Sun, Zhenman Fang, Peipei Zhou, Peichen Pan, and Jason
  Cong.
\newblock Caffeine: Towards uniformed representation and acceleration for deep
  convolutional neural networks.
\newblock {\em IEEE Transactions on Computer-Aided Design of Integrated
  Circuits and Systems}, 2018.

\bibitem{vivado}
Tom Feist.
\newblock Vivado design suite.
\newblock {\em White Paper}, 5:30, 2012.

\bibitem{Quartus}
Intel.
\newblock Intel quartus prime software.

\bibitem{onnx}
ONNX.
\newblock {Open neural network exchange format}, 2018.

\bibitem{hls4mlstatu}
hls4ml project current status.
\newblock \url{https://hls-fpga-machine-learning.github.io/hls4ml/STATUS.html}.
\newblock Accessed: 2019-07-13.

\bibitem{aydonat2017opencl}
Utku Aydonat, Shane O'Connell, Davor Capalija, Andrew~C Ling, and Gordon~R
  Chiu.
\newblock An opencl deep learning accelerator on arria 10.
\newblock In {\em Proceedings of the 2017 ACM/SIGDA International Symposium on
  Field-Programmable Gate Arrays}, pages 55--64. ACM, 2017.

\bibitem{suda2016throughput}
Naveen Suda, Vikas Chandra, Ganesh Dasika, Abinash Mohanty, Yufei Ma, Sarma
  Vrudhula, Jae-sun Seo, and Yu~Cao.
\newblock Throughput-optimized opencl-based fpga accelerator for large-scale
  convolutional neural networks.
\newblock In {\em Proceedings of the 2016 ACM/SIGDA International Symposium on
  Field-Programmable Gate Arrays}, pages 16--25. ACM, 2016.

\bibitem{zhang2015optimizing}
Chen Zhang, Peng Li, Guangyu Sun, Yijin Guan, Bingjun Xiao, and Jason Cong.
\newblock Optimizing fpga-based accelerator design for deep convolutional
  neural networks.
\newblock In {\em Proceedings of the 2015 ACM/SIGDA International Symposium on
  Field-Programmable Gate Arrays}, pages 161--170. ACM, 2015.

\bibitem{ma2016scalable}
Yufei Ma, Naveen Suda, Yu~Cao, Jae-sun Seo, and Sarma Vrudhula.
\newblock Scalable and modularized rtl compilation of convolutional neural
  networks onto fpga.
\newblock In {\em 2016 26th International Conference on Field Programmable
  Logic and Applications (FPL)}, pages 1--8. IEEE, 2016.

\bibitem{pipecnn2019abm}
Dong Wang, Ke~Xu, Qun Jia, and Soheil Ghiasi.
\newblock Abm-spconv: A novel approach to fpga-based acceleration of
  convolutional neural network inference.
\newblock In {\em Proceedings of the 56th Annual Design Automation Conference
  2019}, page~87. ACM, 2019.

\bibitem{haq}
Kuan Wang, Zhijian Liu, Yujun Lin, Ji~Lin, and Song Han.
\newblock Haq: Hardware-aware automated quantization with mixed precision.
\newblock In {\em Proceedings of the IEEE Conference on Computer Vision and
  Pattern Recognition}, pages 8612--8620, 2019.

\bibitem{releq}
Amir Yazdanbakhsh, Ahmed~T Elthakeb, Prannoy Pilligundla, and FatemehSadat
  Mireshghallah~Hadi Esmaeilzadeh.
\newblock Releq: An automatic reinforcement learning approach for deep
  quantization of neural networks.
\newblock {\em arXiv preprint arXiv:1811.01704}, 2018.

\bibitem{a3csurvay}
Ivo Grondman, Lucian Busoniu, Gabriel~AD Lopes, and Robert Babuska.
\newblock A survey of actor-critic reinforcement learning: Standard and natural
  policy gradients.
\newblock {\em IEEE Transactions on Systems, Man, and Cybernetics, Part C
  (Applications and Reviews)}, 42(6):1291--1307, 2012.

\bibitem{zhao2017accelerating}
Ritchie Zhao, Weinan Song, Wentao Zhang, Tianwei Xing, Jeng-Hau Lin, Mani
  Srivastava, Rajesh Gupta, and Zhiru Zhang.
\newblock Accelerating binarized convolutional neural networks with
  software-programmable fpgas.
\newblock In {\em Proceedings of the 2017 ACM/SIGDA International Symposium on
  Field-Programmable Gate Arrays}, pages 15--24. ACM, 2017.

\bibitem{xiao2017exploring}
Qingcheng Xiao, Yun Liang, Liqiang Lu, Shengen Yan, and Yu-Wing Tai.
\newblock Exploring heterogeneous algorithms for accelerating deep
  convolutional neural networks on fpgas.
\newblock In {\em 2017 54th ACM/EDAC/IEEE Design Automation Conference (DAC)},
  pages 1--6. IEEE, 2017.

\bibitem{guide_conv}
Vincent Dumoulin and Francesco Visin.
\newblock A guide to convolution arithmetic for deep learning.
\newblock {\em arXiv preprint arXiv:1603.07285}, 2016.

\bibitem{gajski1983new}
Daniel~D Gajski and Robert~H Kuhn.
\newblock New vlsi tools.
\newblock {\em Computer}, (12):11--14, 1983.

\bibitem{de0nano}
Terasic.
\newblock {DE0-Nano-SoC Kit/Atlas-SoC Kit}.
\newblock \url{de0-nano-soc.terasic.com}, 2019.

\bibitem{de1soc}
Terasic.
\newblock {DE1-SoC Board}.
\newblock \url{de1-soc.terasic.com}, 2019.

\bibitem{nallatech510}
Nallatech.
\newblock {Nallatech 510 Acceleration Board}.
\newblock \url{https://www.bittware.com/fpga/510t/}, 2019.

\bibitem{mnih2016asynchronous}
Volodymyr Mnih, Adria~Puigdomenech Badia, Mehdi Mirza, Alex Graves, Timothy
  Lillicrap, Tim Harley, David Silver, and Koray Kavukcuoglu.
\newblock Asynchronous methods for deep reinforcement learning.
\newblock In {\em International conference on machine learning}, pages
  1928--1937, 2016.

\bibitem{van2007reinforcement}
Hado Van~Hasselt and Marco~A Wiering.
\newblock Reinforcement learning in continuous action spaces.
\newblock In {\em 2007 IEEE International Symposium on Approximate Dynamic
  Programming and Reinforcement Learning}, pages 272--279. IEEE, 2007.

\bibitem{alexnet}
Alex Krizhevsky, Ilya Sutskever, and Geoffrey~E Hinton.
\newblock Imagenet classification with deep convolutional neural networks.
\newblock In {\em Advances in neural information processing systems}, pages
  1097--1105, 2012.

\bibitem{vgg}
Karen Simonyan and Andrew Zisserman.
\newblock Very deep convolutional networks for large-scale image recognition.
\newblock {\em arXiv preprint arXiv:1409.1556}, 2014.

\bibitem{cpu_benchmarking}
Shaohuai Shi, Qiang Wang, Pengfei Xu, and Xiaowen Chu.
\newblock Benchmarking state-of-the-art deep learning software tools.
\newblock In {\em 2016 7th International Conference on Cloud Computing and Big
  Data (CCBD)}, pages 99--104. IEEE, 2016.

\bibitem{qiu2016going}
Jiantao Qiu, Jie Wang, Song Yao, Kaiyuan Guo, Boxun Li, Erjin Zhou, Jincheng
  Yu, Tianqi Tang, Ningyi Xu, Sen Song, et~al.
\newblock Going deeper with embedded fpga platform for convolutional neural
  network.
\newblock In {\em Proceedings of the 2016 ACM/SIGDA International Symposium on
  Field-Programmable Gate Arrays}, pages 26--35. ACM, 2016.

\end{thebibliography}


\end{document}